\documentclass[runningheads]{llncs}

\usepackage{spconf,amsmath,graphicx}
\usepackage{hyperref}
\ninept
\usepackage[section]{placeins}
\usepackage{stfloats}

\title{Optimizing Vision Transformers for Medical Image Segmentation}
\name{ \parbox{\linewidth}{\centering
Qianying Liu\textsuperscript{1}, Chaitanya Kaul\textsuperscript{1}, Jun Wang\textsuperscript{2}, Christos Anagnostopoulos\textsuperscript{1},  Roderick Murray Smith\textsuperscript{1}, *Fani Deligianni\textsuperscript{1} } \thanks{Thanks to EPSRC (EP/W01212X/1), UKRI project 104690 (iCAIRD) and China Scholarship Councils for funding.   
\textsuperscript{1} https://github.com/kathyliu579/CS-Unet } } 

\address{ \textsuperscript{1}School of Computing Science, University of Glasgow \textsuperscript{2}University of Warwick}
\begin{document}

%
\maketitle

\begin{abstract}
For medical image semantic segmentation (MISS), Vision Transformers have emerged as strong alternatives to convolutional neural networks thanks to their inherent ability to capture long-range correlations. However, existing research uses off-the-shelf vision Transformer blocks based on linear projections and feature processing which lack spatial and local context to refine organ boundaries. Furthermore, Transformers do not generalize well on small medical imaging datasets and rely on large-scale pre-training due to limited inductive biases. 
To address these problems, we demonstrate the design of a compact and accurate Transformer network for MISS, CS-Unet, which introduces convolutions in a multi-stage design for hierarchically enhancing spatial and local modeling ability of Transformers. This is mainly achieved by our well-designed Convolutional Swin Transformer (CST) block which merges convolutions with Multi-Head Self-Attention and Feed-Forward Networks for providing inherent localized spatial context and inductive biases. Experiments demonstrate CS-Unet without pre-training outperforms other counterparts by large margins on multi-organ and cardiac datasets with fewer parameters and achieves state-of-the-art performance. Our code is available at Github\textsuperscript{1}.

\end{abstract}

\section{Introduction}
\label{sec:intro}

Medical image semantic segmentation (MISS), which classifies image pixels with semantic organ labels (e.g. Kidney and Liver) for various imaging modalities, is considered as one of the most fundamental problems in medical imaging. However, compared to natural scene images, MISS requires overcoming more challenges to create robust models. For instance, common benchmark datasets in MISS suffer from large deformation of organs under different image acquisition processes. In addition, shortage of costly pixel-level annotations is another problem leading to a performance gap. To achieve efficient and effective segmentation, models are not only required to have a better understanding of their local semantic features to capture more subtle organ structures, but also of global feature dependencies to capture the relationships among multiple organs. 

UNet \cite{Ronneberger2015} and its variants \cite{Zhou2018}\cite{Huang2020}\cite{Schlemper2019}\cite{kaul2019focusnet}\cite{kaul2021focusnet++} with Convolutional Neural Networks (CNNs) as the backbone have found huge success in MISS as they are good at modelling local attributes inside their receptive field. However, the inherent locality of convolution operations restricts their ability to model long-range semantic dependencies within the image, and as a result the challenging boundaries of the whole organ may not be effectively segmented. Attention mechanisms alleviate this issue, but these tend to be 'single head' mechanisms that only calculate pixel-level similarities, and not 'multi head' with the ability to capture patch-level patterns.

For alleviating the inherent flaws of CNNs, there's a recent shift in the choice of architectures from CNNs to Vision Transformers (ViTs) due to their ability to model long range semantic attributes among input tokens (embeddings of image patches) via a linearly projected Multi-Head Self-Attention (MHSA) operation and a Feed-Forward Network (FFN). Most early works \cite{Chen2021Transunet}\cite{zhang2021}\cite{Chen2021a} treat CNNs as a backbone and exploit the Transformer's desirable characteristics in their encoder. They tend to have high complexity as they stack bulky Transformer blocks on top of convolutional feature extractors (large pretrained CNNs, e.g. ResNet). Recent research \cite{valanarasu2021}\cite{Karimi2021}\cite{Cao2021}\cite{Lin2022}\cite{Huang2021}\cite{Zhou2021}\cite{Wang2022}\cite{Tragakis2022} has moved towards using Transformers as the main stem for building the entire segmentation architecture. Swin-UNet \cite{Cao2021} is regarded as the first pure Transformer model. It keeps the familiar U-shape and adds hierarchical feature extraction using shifted windows proposed by the Swin Transformer \cite{liu2021}. This drastically reduces the quadratic complexity of traditional self-attention while achieving better performance. 

\begin{figure*}[h]
\begin{center}
\includegraphics[height=3.6cm,width=17.5cm]{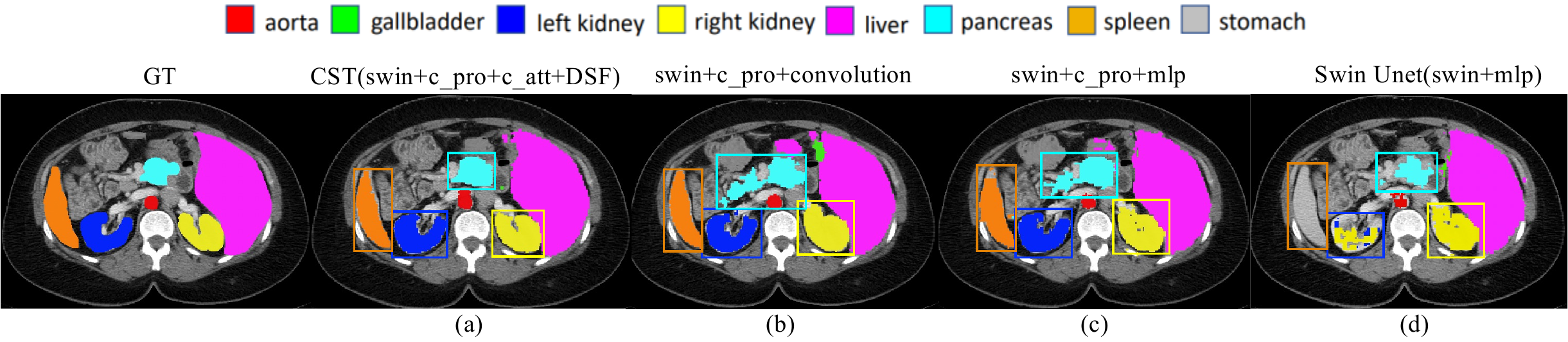}
\end{center}
\vspace{-13pt}
\caption{Visualization of segmentation results of different methods trained from scratch on Synapse dataset.}
\vspace{-13pt}
\label{fig:compare}
\end{figure*}

However, most of these Transformers for MISS use off-the-shelf Transformer blocks from Computer Vision community and only model and extract linear semantic relations via MHSA and FFN, leading to the challenge of precisely delineating organ boundaries due to the lack of spatial and local information as shown in Figure \ref{fig:compare}.(d), although showing small influence on detection and classification tasks. Besides, these methods require a large dataset to compensate the lack of inductive biases such as translation equivariance \cite{dosovitskiy2020image}, which may be defected or even lost when fine-tuning on downstream tasks, showing less robustness on small datasets.

Keeping the current state of the literature in mind, our paper highlights issues that today's Transformers in MISS face, followed by our contribution that helps alleviate those drawbacks. Most current Transformers are bulky and rely on pre-training weights from classical vision tasks to be adapted for MISS. To the best of our knowledge, no existing study explores the effects of adding spatial locality inside Transformer blocks via convolutions for medical imaging. 
To this end, we first propose an empirical analysis to show the need for spatial locality in pure Transformer based MISS. Our insights show the effects of introducing convolutions to Transformer blocks and multi-stage design of networks on segmentation performance. We call the final model resulting from our experiments, Convolutional Swin-Unet (CS-Unet), which is based on purely convolutional Transformer blocks created to make Transformers model local information better, segment organ boundaries more accurately, while maintaining a low computational complexity. Experiments on CT and MRI datasets show CS-Unet (24M parameters) trained from scratch outperforms pre-trained Swin-Unet (27M) on ImageNet by around 3$\%$ dice score, achieving state-of-the-art performance.

\section{Method}

Most Transformer based methods in MISS, i.e., encoder-decoder models with a standard U-shape, use a standard Transfomer block containing linear projections and linear FFNs, which are essentially MLPs, to process the data. Hence, to create effective image representations using such a regime requires huge amounts of data for training, as they lack local spatial information. 

The first pure-Transformer based MISS model is the Swin-Unet \cite{Cao2021} which adopts Swin Transformer blocks \cite{liu2021} to add locality information to Transformers. The data representation created here is still inherently linear as this block contains linear projections and feature processing. 
Figure \ref{fig:compare} shows segmentation visualizations for the Synapse dataset. Swin-Unet trained from a random weight initialization (Figure \ref{fig:compare}.(d)) does not perform well. It fails to detect the spleen and misclassifies the left kidney as the right.


Next, we add convolutional projections to this Swin Transformer block structure. The projections follow the methodology proposed in \cite{Wu2021} where tokens are first shaped into a 2D token map, then processed by a depth-wise separable convolution with kernel size $s$ implemented by: {\tt Depth-wise Conv $\rightarrow$ BatchNorm $\rightarrow$ Point-wise Conv}. Finally, the tokens are flattened into 1D token input $x_{i}^{q/k/v}$ for Q/K/V matrices. It can be formulated as: $x_{i}^\mathrm{q/k/v} =Flatten(Conv(Reshape2D(x_{i}),s)$.
Figure \ref{fig:compare}.(c) shows outputs of the resultant Unet trained with this block. It visually demonstrates how spatial locality is essential for low level pixel labelling tasks. It can be seen that although the convolutional projection alleviates a lot of the problems posed by the linearity of Swin-Unet, there are still severe over-segmentations on pancreas and liver and extremely rough boundaries of right kidney.


Following this, when a 3x3 convolution is used for FFNs instead of MLPs to introduce more spatial context, we see the full effects of adding complete spatial locality to Transformers through the boundaries of the left and right kidneys and spleen becoming greatly refined. The over-segmentation problem of the pancreas however gets worse (as shown in Figure \ref{fig:compare}.(b)). This is due to the limited receptive field not modeling the whole boundary of big organs effectively. 
\vspace{-0.3cm}
\subsection{Convolutional Swin Transformer (CST) Layer} 

We propose a CST layer to fully explore spatial modeling ability of convolutions in MHSA and FFN. First, we propose a novel (shifted) window based convolutional multi-head self attention ((S)W-CMSA) to extract hierarchical semantic features while reducing computational costs, by combining a shifted windows mechanism and convolutional projection. Then, we replace the MLP with our novel depthwise separable feed-forward (DSF) module. From Figure \ref{fig:compare}.(a), we see the Transformer model based on CST handles challenging organ boundaries more efficiently. The CST layer is formulated as:

\begin{align}
\label{att}
\hat{z}^\mathrm{l} &= W-CMSA(LN(z^\mathrm{l-1}))+z^\mathrm{l-1}, \\
z^\mathrm{l} &=DSF(\hat{z}^\mathrm{l})+\hat{z}^\mathrm{l}, \\
\hat{z}^\mathrm{l+1} &=SW-CMSA(LN(z^\mathrm{l}))+z^\mathrm{l},  \\
\hat{z}\mathrm{l+1} &=DSF(\hat{z}^\mathrm{l+1})+\hat{z}^\mathrm{l+1}
\vspace{-13pt}
\end{align}

where $\hat{z}^{l}$ and $z^{l}$ denote the outputs of (S)W-CMSA module and DSF of the $l$-th block, respectively.

{\bf (Shifted) Window based convolutional multi-head self attention}

\begin{figure}[h]
\begin{center}
\includegraphics[height=3.6cm,width=8.7cm]{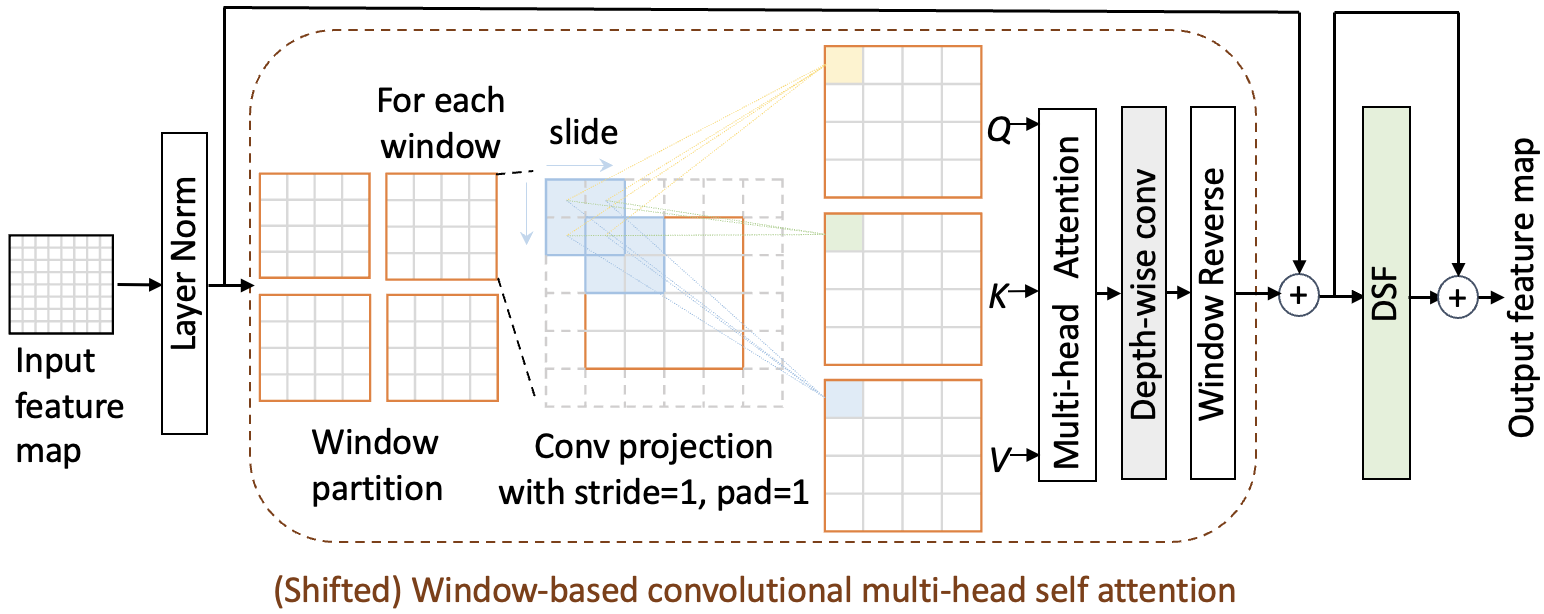}
\end{center}
\vspace{-13pt}
\caption{Convolutional Swin Transformer (CST) Block.}
\label{fig:CSTblock}
\vspace{-8pt}
\end{figure}

As shown in Figure \ref{fig:CSTblock}, once tokens enter (S)W-CMSA, they are reshaped into a 2D token map, and partitioned into windows. For each window, we use three depth-wise convolutions with kernel size $s$ of 3×3, padding of 1 and stride of 1 to create our Q, K and V vectors via: $Flatten(DepthConv(Window(Reshape(x_{i})),s)$.


CST is different from \cite{Wu2021} as we create a projection based on windows rather than the whole image, leading to more refined local features as now the kernels learnt on each window are different. In order to better adapt to medical images with smaller data volumes, point-wise convolutions are removed to avoid over-fitting. Furthermore, we replace Batch Normalization with Layer Normalization (LN), providing a performance boost. The token vectors are fed to MHSA as:

\begin{equation}
\begin{aligned}
MHSA(x_{i}^\mathrm{q}, x_{i}^\mathrm{k}, x_{i}^\mathrm{v}) =SoftMax &\left( \frac{x_{i}^\mathrm{q}
(x_{i}^\mathrm{k})^\mathrm{T}}{\sqrt{d}} +B \right)x_{i}^\mathrm{v}
\end{aligned}
\label{att2}
\vspace{-5pt}
\end{equation}

Here $d$ represents the dimension of the query and key. The values in $B$ are the bias. 

Then, we replace the linear layer and feed the attention output to a 3×3 depth-wise convolution for fine-tuning for more spatial information.
We follow this by reversing the windows to 2D token maps, resulting in more robust estimations compared to Swin Transformer \cite{liu2021} removing our dependence on positional encoding.

{\bf Depthwise separable feed-forward (DSF) module} After computing (S)W-CMSA, the feature maps are fed into a FFN. Existing Transformers implement this module as an MLP: {\tt LN,d $\rightarrow$ Linear,4×d $\rightarrow$ GELU $\rightarrow$ Linear,d $\rightarrow$ RC}. The $d$ denotes the number of channels of a reshaped feature map and $RC$ denotes the residual connection.
We propose a DSF module as a choice of FFN which provides adding spatial context. We use three depth-wise convolutions instead of two linear layers for utilizing the features between channels, $C$. In addition, we found that adding LN after convolution gives better segmentation results. The DSF is implemented as: {\tt 7x7 Depth-wise Conv,d $\rightarrow$ LN,d $\rightarrow$ Point-wise Conv,4×d $\rightarrow$ GELU $\rightarrow$ Point-wise Conv,d $\rightarrow$ RC}.

\subsection{Overall Structure Design}  
\begin{figure}[h]
\begin{center}
\includegraphics[width=0.45\textwidth]{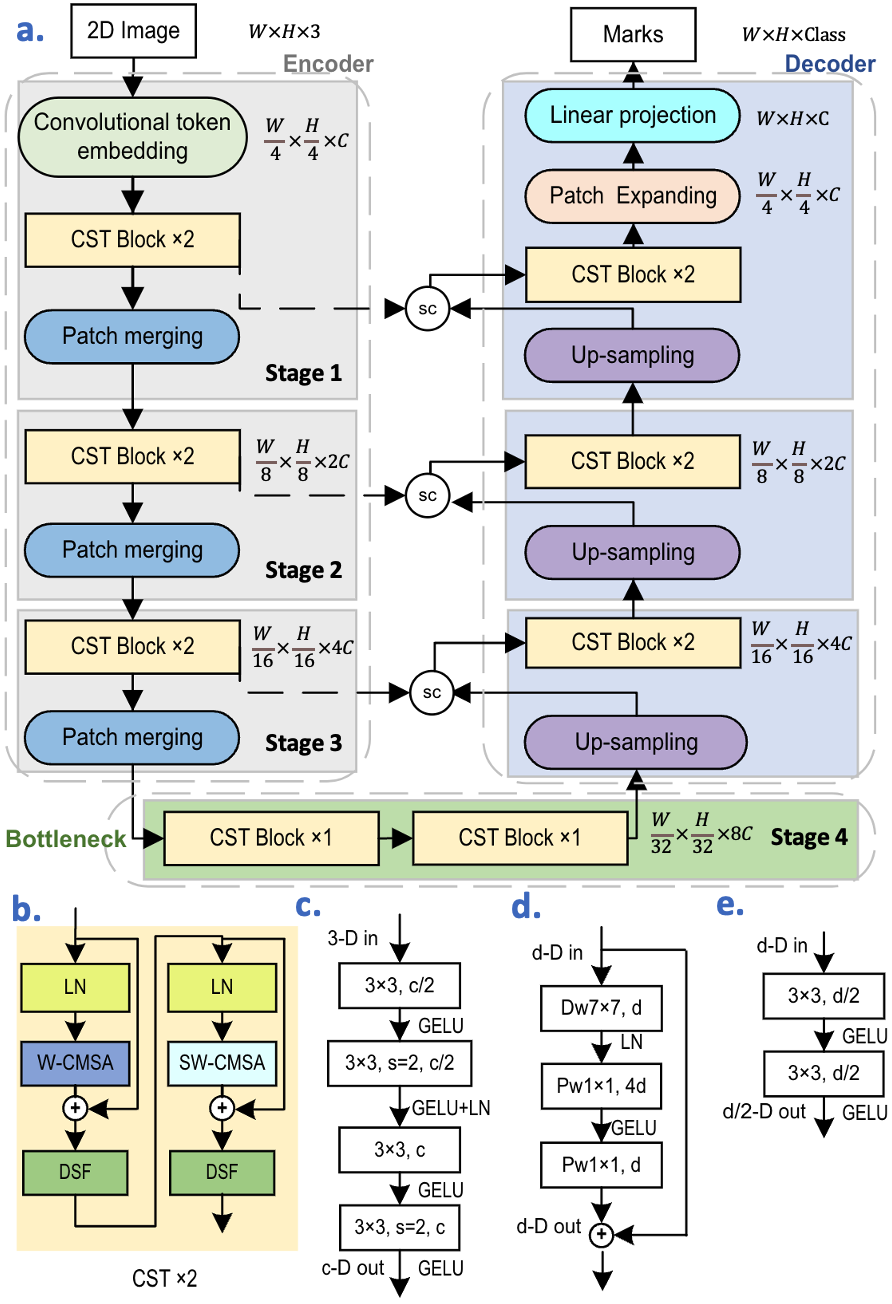}
\end{center}
\vspace{-13pt}
\caption{ (a) Overall architecture of CS-Unet, (b) one CST layer, (c) convolutional token embedding, (d) DSF and (e) skip convolutions. $d$ is the current number of channels, $c$ is an arbitrary dimension.}
\label{fig:CsU}
\vspace{-15pt}
\end{figure}

CS-Unet keeps a symmetrical UNet shape. The input of our model is a 2D image slice with the resolution of $H\times W\times 3$ sampled from a 3D volume of images. $H$, $W$ and $3$ denote the height, width and number of channels of each input. The input images on entering the encoder are passed through the convolutional token embedding to create a sequence embedding on overlapping patches of the image, following which CST and patch merging layers are applied. Extracted features are then processed by the model's bottleneck that consists of two CST blocks. A symmetrical decoder then creates the final segmentation marks. In addition, skip convolution (SC) modules are added between corresponding feature pyramids of the encoder and decoder to compensate for the missing information caused by down-sampling. The overall architecture of the proposed CS-Unet is presented in Figure \ref{fig:CsU}.
 
\subsection{Encoder}
The input image is first passed through the convolutional token embedding layer to create a sequence embedding with the resolution of $\frac H4\frac W4\times C$ ($C$ = 96 in experiments). This embedding is fed to three main CST layers and a patch merging module which downsamples the image and doubles the number of channels. For example, at the first patch merging module, an input with size $\frac H4\times \frac W4\times C$ is divided into four parts and concatenated along the $C$ dimension to create a feature map of size $\frac H8\times \frac W8\times 4C$. Then a linear layer is applied to this map to reduce the $C$ dimension by a factor of 2.  

{\bf Convolutional Token Embedding layer} 
Existing models use a linear layer to split images into non-overlapping patches and reduce the size of the image drastically, e.g. by $75\%$, while increasing the channel dimension $C$. However, as the images' highest resolution is $ H\times W$ at the encoder, using a linear layer to compress these features not only loses high-quality spatial and local information, but also increases model size. Our embedding layer, is implemented as four convolutions with overlapping patches to compress features in stages, helping to introduce more spatial dependency between, and inside the patches, while greatly reducing the parameters (by 6M. See Ablation \ref{tab:ablation}, Method 1). Specifically, this layer is implemented as follows: {\tt 3x3 s=1 Conv,d/2 $\rightarrow$ GELU $\rightarrow$ 3x3 s=2 Conv,d/2 $\rightarrow$ GELU+LN $\rightarrow$ 3x3 s=1 Conv,d $\rightarrow$ GELU $\rightarrow$ 3x3 s=2 Conv,d $\rightarrow$ GELU}. Here, $s$ is stride, the input dimension is 3, and $d=C$. In the end, 2D reshaped token maps with resolution $\frac H4\times \frac W4\times C$ are outputted.

{\bf Bottleneck }
The bottleneck contains two CST blocks, based on W-CMSA. The feature map size here remains unchanged.


\subsection{Decoder}
Our decoder is symmetric to the encoder. Feature representation is created by enlarging the feature volume through a convolutional up-sampling module and then passing it through a SC module to compensate for the information lost due to down-sampling. A CST layer then provides spatial context to the upsampled features. After repeating the above process three times, the features are fed into the patch expansion layer which up-samples by $4\times$, followed by a linear projection to fine tune the final segmentation prediction. Specifically, convolutional up-sampling module employs strided deconvolution to $2\times$ up-sample feature maps and halves the channel dimension as: {\tt LN,d $\rightarrow$ 2x2 s=2 ConvTranspose,d/2 $\rightarrow$ GELU}. 

{\bf Skip Convolutions (SC) module }
The outputs of high-resolution feature maps created from up-sampling are concatenated with shallow feature representations from the encoder, and then merged by a SC module. It further enriches both spatial and fine-grained information, while compensating for the missing information caused by down-sampling. It is implemented as :{\tt 3x3 s=1 Conv,d/2 $\rightarrow$ GELU $\rightarrow$ 3x3 s=1 Conv,d/2 $\rightarrow$ GELU}.

\section{Experiments}
\begin{table*}
\begin{minipage}[c]{0.65\textwidth}
\centering
\scalebox{0.81}{
\begin{tabular}{|c|c|c|c c c c c c c c|}
\hline
Methods & DSC & HD & Aorta & Gallb & Kid$\_$L & Kid$\_$R& Liver& Pancr & Spleen& Stom\\
\hline
R50 UNet \cite{Chen2021Transunet} & 74.68  & 36.87 & 84.18 & 62.84 & 79.19 &71.29 &93.35 &48.23 &84.41 &73.92 \\
R50 AttnUNet \cite{Chen2021Transunet} & 75.57 &36.97&55.92 &	63.91&79.20&72.71&93.56&49.37	&87.19	&74.95 \\
UNet \cite{Ronneberger2015}  & 76.85 & 39.70& \underline{89.07} & \underline {69.72}& 77.77& 68.60& 93.43& 53.98& 86.67& 75.58 \\
AttnUNet \cite{Schlemper2019} & 77.77& 36.02& {\bf89.55}   & 	68.88& 77.98& 71.11& 93.57	& 58.04	& 87.30	& 75.75 \\
\hline
R50 ViT \cite{Chen2021Transunet} & 71.29 & 	32.87& 	73.73  & 55.13& 	75.80& 	72.20& 	91.51& 	45.99& 	81.99& 	73.95 \\
TransUnet \cite{Chen2021Transunet} & 77.48& 31.69& 87.23 & 63.13& 	81.87& 	77.02& 	94.08& 	55.86	& 85.08& 	75.62 \\
Swin-Unet \cite{Cao2021} & \underline{79.13} & 	{\bf 21.55}& 	85.47& 66.53& 	\underline{83.28}& 	{\bf 79.61}&  \underline{94.29}& 	56.58& 	 \underline{90.66} & 	\underline{76.60} \\
MT-UNet \cite{Wang2022} & 78.59& 	\underline{26.59}& 	87.92 &  64.99& 	81.47& 	77.29& 	93.06& 	\underline{59.46}& 	87.75& 	{\bf76.81}\\
\hline
Ours &  \bf{82.21} &  27.02 & 88.40& 	{\bf 72.59}	& {\bf 85.28}& 	\underline{79.52}& 	{\bf94.35}& 	{\bf 70.12}& {\bf91.06}& 	 75.72 \\
\hline
\end{tabular}}
\caption{Comparison with different models on Synapse. Gallbladder, left Kidney, right Kidney, Pancreas and Stomach are abbreviated as Gallb, Kid$\_$L, Kid$\_$R, Pancr and Stom.}
\label{tab:synapse}
\end{minipage}
\begin{minipage}[c]{0.36\textwidth}
\centering
\scalebox{0.85}{
\begin{tabular}{|c|c|c c c|}
\hline
Methods & DSC & RV & Myo & LV \\
\hline
R50 UNet  & 87.60 & 84.62 & 84.52 & 93.68 \\
R50 AttnUNet & 86.90 & 83.27 & 84.33 & 93.53 \\
R50 ViT  & 86.19 & 82.51 & 83.01 & 93.05 \\
TransUnet  & 89.71 & \underline{86.67} & 87.27 & 95.18 \\
Swin-Unet & 88.07 & 85.77 & 84.42 & 94.03 \\
MT-UNet & \underline{90.43} & 86.64 & \underline{89.04} & {\bf 95.62} \\
\hline
Ours & {\bf 91.37} & {\bf 89.20} & {\bf 89.47} & \underline{95.42} \\
\hline
\end{tabular}
 }
\caption{Experimental results of ACDC.}
\label{tab:acdc}
\vspace{24pt}

\end{minipage}
\end{table*}

We use two publicly available datasets to benchmark our method.

{\bf Synapse multiorgan segmentation (Synapse):}
This dataset \cite{landman2015miccai} contains abdominal CT scans from 30 subjects. Following \cite{Chen2021Transunet}, 18 cases (2212 axial slices) are extracted for training, while other 12 cases are used for testing. We report the model performance evaluated with the average Dice score Coefficient (DSC) and average Hausdorff Distance (HD) on eight abdominal organs. 

{\bf Automatic Cardiac Diagnosis Challenge (ACDC):}
ACDC \cite{bernard2018deep} contains MRI images from 100 patients, with right ventricle (RV), left ventricle (LV) and myocardium (MYO) to be segmented. Using data splits proposed in \cite{Wang2022}, the dataset is split into 70 (1930 axial slices), 10 and 20 for training, validation and testing, respectively. Evaluation metrics used are average DSC ($\%$) and HD (mm).

\subsection{Implementation details}
We train our models on a single Nvidia RTX3090 GPU with 24GB memory. We use flipping and rotation augmentations on the training data. The input image size is 224×224. Pre-trained weights are used for other methods if provided, while our model is trained from scratch for 300 epochs from a randomly initialized set of weights. A batch size of 24 and a combination of cross entropy and dice loss are used. Our model is optimized by AdamW \cite{loshchilov2018decoupled} with a weight decay of 5E-4 for both datasets. The learning rates for Synapse and ACDC are 1e-3 and 5e-3, respectively. We start with a 10-epoch linear warmup. Layer Scale \cite{Touvron2021} of initial value 1e-6 is applied.

\subsection{Experimental Results}


As shown in \autoref{tab:synapse} and \autoref{tab:acdc}, our model consistently surpasses a variety of convolution-based and Transformer-based methods. CS-Unet outperforms Swin-Unet by 3.08\% and 3.3\% DSC on Synapse and ACDC, respectively. In addition, our method gets the highest DSC for five and two organs of Synapse and ACDC respectively, especially providing large boosts for challenging organs like gallbladder, pancreas and RV. Overall, compared to pretrained Swin-Unet (27 M) and TransUnet (96 M), CS-Unet achieves the best performance without pretraining while being lightweight (24 M) via introducing more local perception and inductive bias.

\begin{figure}[h]
\begin{center}
\includegraphics[height=4.3cm,width=8.7cm]{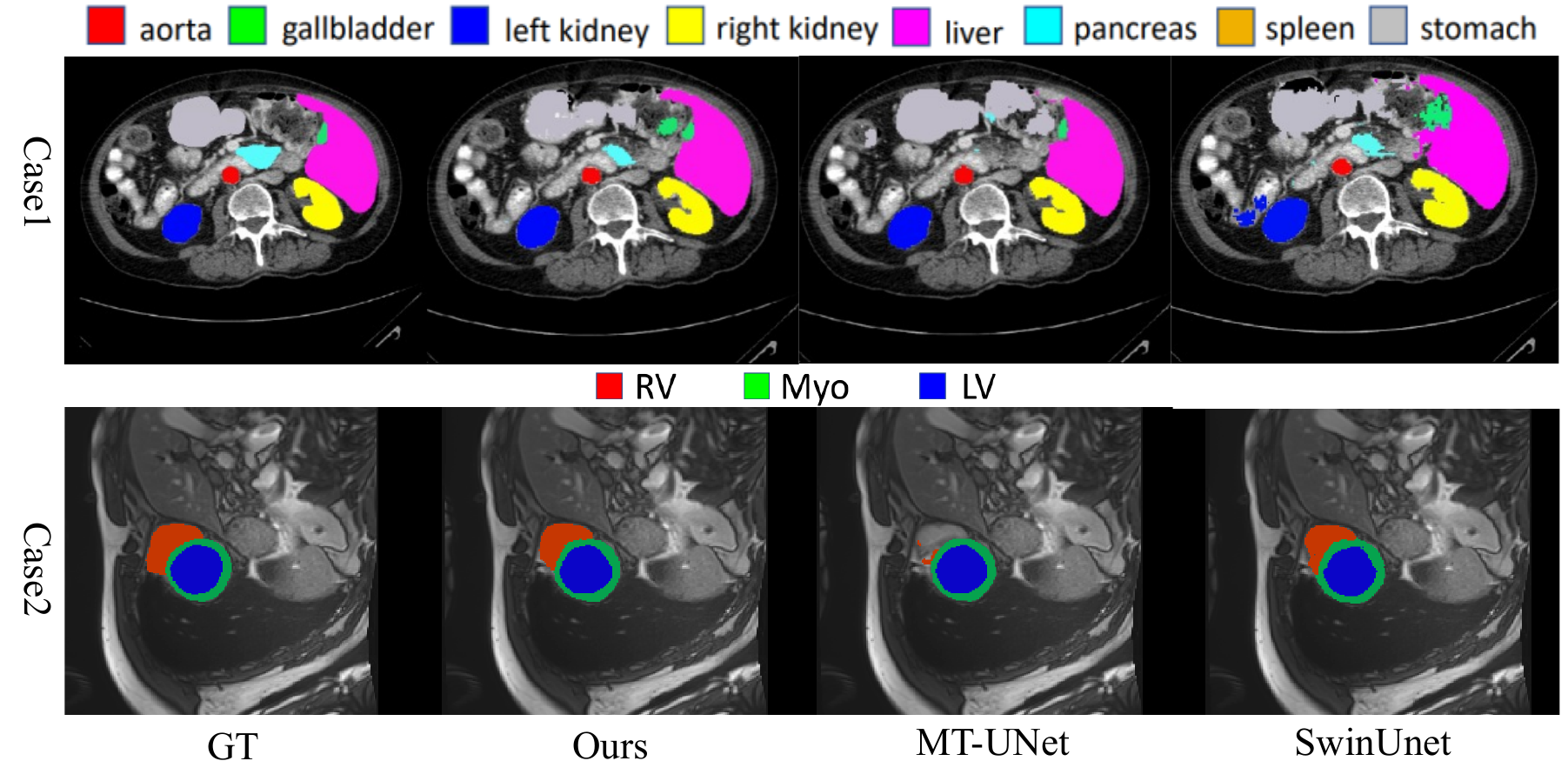}
\end{center}
\vspace{-13pt}
\caption{Visualization of segmentation results on two datasets.}
\label{fig:Visual}
\vspace{-6pt}
\end{figure}

Figure \ref{fig:Visual} visualizes segmentation results. In case 1, our method has overwhelming advantage on segmenting the pancreas, stomach and liver. CS-Unet is also more discriminative on the complex shape of RV than other Transformer-based models in case 2 due to its better ability of spatial context modelling.

\subsection{Ablation study}
We explore the influence of proposed modules on the performance on Synapse as shown in Table \ref{tab:ablation}. The Swin-Unet trained from scratch is treated as the baseline (method 0) which cannot adapt to small datasets. 
Adding convolutional token embedding (method 1) and convolutional projections (method 2), we observe large improvements of $8\%$ and $9\%$ on DSC which is competitive with pre-trained Swin-Unet emphasizing the importance of adding local modeling ability to Transformers. 
Removing the position embedding in early stages and using a convolution instead of a linear layer to fine-tune the attention computation (method 3) leads to a slight increase in performance and parameter reduction. 
Method 4 combines the CST block with the DSF module leading to an improved DSC and HD without extra parameters. 
After utilizing convolutional up-sampling and feature fusion module, SC, for merging information during skip connection, our best performing model method 6 achieves 3.17\% improvements on DSC. Comparison with method 5, shows that fully convolutional pure Transformers can track the position of pixels better without needing an extra positional embedding, and that spatial feature extraction, is in fact, a necessity for Transformers.

\begin{table}[h]
\begin{center}
\scalebox{0.78}{
\begin{tabular}{|c|c|c|c c c c c c|c|}
\hline
 Methods& DSC & HD & Emb & Proj & Pos & Att & DSF & SC & $\#$param \\

\hline
0 (Base) &60.80 &54.35 & & &$\surd$ & & & & 27.15\\
1 & 68.57& 51.02& $\surd$ & & $\surd$& & & &21.55\\

2 & 77.47 & \bf{18.54}& $\surd$ & $\surd$&$\surd$ & & & &21.55\\
3 & 78.32 & 25.43 & $\surd$ & $\surd$ & $\times$  & $\surd$ & &  & 19.63 \\
4 & 79.04 & 22.96 & $\surd$ & $\surd$ & $\times$  & $\surd$ & $\surd$&  & 19.84 \\
5 & 81.93& 24.59 &$\surd$ &$\surd$ &$\surd$  &$\surd$ &$\surd$  &$\surd$ & 24.68\\
    \hline
6 & \bf{82.21} &  27.02 & $\surd$ & $\surd$ & $\times$ & $\surd$ & $\surd$& $\surd$& 24.68 \\
\hline
\end{tabular}
}
\end{center}
\vspace{-13pt}
\caption{Ablation study on modules used in CS-Unet on Synapse.}
\label{tab:ablation}
\vspace{-8pt}
\end{table}

\section{Conclusions}

In this work, we presented the effects of introducing convolutions to Transformer blocks and to a multi-stage Transformer network to alleviate limitations of non-locality and need for extensive pre-training that Transformers in MISS face. Extensive experiments demonstrated that merging Convolutions with MHSAs and FFNs to create our CST layer, provided inherent local context inside Transformer blocks. Based on CST, our compact, accurate and pure Transformer architecture, CS-Unet, achieved superior performance without pretraining while maintaining less parameters.

\bibliographystyle{IEEEbib}
\bibliography{strings,refs}

\begin{thebibliography}{10}

\bibitem{Ronneberger2015}
Olaf Ronneberger, Philipp Fischer, and Thomas Brox,
\newblock ``U-net: Convolutional networks for biomedical image segmentation,''
\newblock in {\em International Conference on Medical image computing and
  computer-assisted intervention}. Springer, 2015, pp. 234--241.

\bibitem{Zhou2018}
Zongwei Zhou, Md~Mahfuzur Rahman~Siddiquee, Nima Tajbakhsh, and Jianming Liang,
\newblock ``Unet++: A nested u-net architecture for medical image
  segmentation,''
\newblock in {\em Deep learning in medical image analysis and multimodal
  learning for clinical decision support}, pp. 3--11. Springer, 2018.

\bibitem{Huang2020}
Huimin Huang, Lanfen Lin, Ruofeng Tong, Hongjie Hu, Qiaowei Zhang, Yutaro
  Iwamoto, Xianhua Han, Yen-Wei Chen, and Jian Wu,
\newblock ``Unet 3+: A full-scale connected unet for medical image
  segmentation,''
\newblock in {\em ICASSP 2020-2020 IEEE International Conference on Acoustics,
  Speech and Signal Processing (ICASSP)}. IEEE, 2020, pp. 1055--1059.

\bibitem{Schlemper2019}
Jo~Schlemper, Ozan Oktay, Michiel Schaap, Mattias Heinrich, Bernhard Kainz, Ben
  Glocker, and Daniel Rueckert,
\newblock ``Attention gated networks: Learning to leverage salient regions in
  medical images,''
\newblock {\em Medical image analysis}, vol. 53, pp. 197--207, 2019.

\bibitem{kaul2019focusnet}
Chaitanya Kaul, Suresh Manandhar, and Nick Pears,
\newblock ``Focusnet: An attention-based fully convolutional network for
  medical image segmentation,''
\newblock in {\em 2019 IEEE 16th international symposium on biomedical imaging
  (ISBI 2019)}. IEEE, 2019, pp. 455--458.

\bibitem{kaul2021focusnet++}
Chaitanya Kaul, Nick Pears, Hang Dai, Roderick Murray-Smith, and Suresh
  Manandhar,
\newblock ``Focusnet++: Attentive aggregated transformations for efficient and
  accurate medical image segmentation,''
\newblock in {\em 2021 IEEE 18th International Symposium on Biomedical Imaging
  (ISBI)}. IEEE, 2021, pp. 1042--1046.

\bibitem{Chen2021Transunet}
Jieneng Chen, Yongyi Lu, Qihang Yu, Xiangde Luo, Ehsan Adeli, Yan Wang, Le~Lu,
  Alan~L Yuille, and Yuyin Zhou,
\newblock ``Transunet: Transformers make strong encoders for medical image
  segmentation,''
\newblock {\em arXiv preprint arXiv:2102.04306}, 2021.

\bibitem{zhang2021}
Yundong Zhang, Huiye Liu, and Qiang Hu,
\newblock ``Transfuse: Fusing transformers and cnns for medical image
  segmentation,''
\newblock in {\em International Conference on Medical Image Computing and
  Computer-Assisted Intervention}. Springer, 2021, pp. 14--24.

\bibitem{Chen2021a}
Bingzhi Chen, Yishu Liu, Zheng Zhang, Guangming Lu, and David Zhang,
\newblock ``Transattunet: Multi-level attention-guided u-net with transformer
  for medical image segmentation,''
\newblock {\em arXiv preprint arXiv:2107.05274}, 2021.

\bibitem{valanarasu2021}
Jeya Maria~Jose Valanarasu, Poojan Oza, Ilker Hacihaliloglu, and Vishal~M
  Patel,
\newblock ``Medical transformer: Gated axial-attention for medical image
  segmentation,''
\newblock in {\em International Conference on Medical Image Computing and
  Computer-Assisted Intervention}. Springer, 2021, pp. 36--46.

\bibitem{Karimi2021}
Davood Karimi, Serge~Didenko Vasylechko, and Ali Gholipour,
\newblock ``Convolution-free medical image segmentation using transformers,''
\newblock in {\em International Conference on Medical Image Computing and
  Computer-Assisted Intervention}. Springer, 2021, pp. 78--88.

\bibitem{Cao2021}
Hu~Cao, Yueyue Wang, Joy Chen, Dongsheng Jiang, Xiaopeng Zhang, Qi~Tian, and
  Manning Wang,
\newblock ``Swin-unet: Unet-like pure transformer for medical image
  segmentation,''
\newblock {\em arXiv preprint arXiv:2105.05537}, 2021.

\bibitem{Lin2022}
Ailiang Lin, Bingzhi Chen, Jiayu Xu, Zheng Zhang, Guangming Lu, and David
  Zhang,
\newblock ``Ds-transunet: Dual swin transformer u-net for medical image
  segmentation,''
\newblock {\em IEEE Transactions on Instrumentation and Measurement}, 2022.

\bibitem{Huang2021}
Xiaohong Huang, Zhifang Deng, Dandan Li, and Xueguang Yuan,
\newblock ``Missformer: An effective medical image segmentation transformer,''
\newblock {\em arXiv preprint arXiv:2109.07162}, 2021.

\bibitem{Zhou2021}
Hong-Yu Zhou, Jiansen Guo, Yinghao Zhang, Lequan Yu, Liansheng Wang, and Yizhou
  Yu,
\newblock ``nnformer: Interleaved transformer for volumetric segmentation,''
\newblock {\em arXiv preprint arXiv:2109.03201}, 2021.

\bibitem{Wang2022}
Hongyi Wang, Shiao Xie, Lanfen Lin, Yutaro Iwamoto, Xian-Hua Han, Yen-Wei Chen,
  and Ruofeng Tong,
\newblock ``Mixed transformer u-net for medical image segmentation,''
\newblock in {\em ICASSP 2022-2022 IEEE International Conference on Acoustics,
  Speech and Signal Processing (ICASSP)}. IEEE, 2022, pp. 2390--2394.

\bibitem{Tragakis2022}
Athanasios Tragakis, Chaitanya Kaul, and Husmeier~Dirk Murray-Smith~Roderick,
\newblock ``The fully convolutional transformer for medical image
  segmentation,''
\newblock in {\em https://arxiv.org/abs/2206.00566}, 2022.

\bibitem{liu2021}
Ze~Liu, Yutong Lin, Yue Cao, Han Hu, Yixuan Wei, Zheng Zhang, Stephen Lin, and
  Baining Guo,
\newblock ``Swin transformer: Hierarchical vision transformer using shifted
  windows,''
\newblock in {\em Proceedings of the IEEE/CVF International Conference on
  Computer Vision}, 2021, pp. 10012--10022.

\bibitem{dosovitskiy2020image}
Alexey Dosovitskiy, Lucas Beyer, Alexander Kolesnikov, Dirk Weissenborn,
  Xiaohua Zhai, Thomas Unterthiner, Mostafa Dehghani, Matthias Minderer, Georg
  Heigold, Sylvain Gelly, et~al.,
\newblock ``An image is worth 16x16 words: Transformers for image recognition
  at scale,''
\newblock in {\em International Conference on Learning Representations}, 2020.

\bibitem{Wu2021}
Haiping Wu, Bin Xiao, Noel Codella, Mengchen Liu, Xiyang Dai, Lu~Yuan, and Lei
  Zhang,
\newblock ``Cvt: Introducing convolutions to vision transformers,''
\newblock in {\em Proceedings of the IEEE/CVF International Conference on
  Computer Vision}, 2021, pp. 22--31.

\bibitem{landman2015miccai}
Bennett Landman, Zhoubing Xu, J~Igelsias, Martin Styner, T~Langerak, and Arno
  Klein,
\newblock ``Miccai multi-atlas labeling beyond the cranial vault--workshop and
  challenge,''
\newblock in {\em Proc. MICCAI Multi-Atlas Labeling Beyond Cranial
  Vault—Workshop Challenge}, 2015, vol.~5, p.~12.

\bibitem{bernard2018deep}
Olivier Bernard, Alain Lalande, Clement Zotti, Frederick Cervenansky, Xin Yang,
  Pheng-Ann Heng, Irem Cetin, Karim Lekadir, Oscar Camara, Miguel
  Angel~Gonzalez Ballester, et~al.,
\newblock ``Deep learning techniques for automatic mri cardiac multi-structures
  segmentation and diagnosis: is the problem solved?,''
\newblock {\em IEEE transactions on medical imaging}, vol. 37, no. 11, pp.
  2514--2525, 2018.

\bibitem{loshchilov2018decoupled}
Ilya Loshchilov and Frank Hutter,
\newblock ``Decoupled weight decay regularization,''
\newblock in {\em International Conference on Learning Representations}, 2019.

\bibitem{Touvron2021}
Hugo Touvron, Matthieu Cord, Alexandre Sablayrolles, Gabriel Synnaeve, and
  Herv{\'e} J{\'e}gou,
\newblock ``Going deeper with image transformers,''
\newblock in {\em Proceedings of the IEEE/CVF International Conference on
  Computer Vision}, 2021, pp. 32--42.

\end{thebibliography}

\end{document}